\documentclass[pdflatex,sn-mathphys-num]{sn-jnl}% Math and Physical Sciences Numbered Reference Style 
\usepackage{graphicx}%
\usepackage{multirow}%
\usepackage{amsmath,amssymb,amsfonts}%
\usepackage{amsthm}%
\usepackage{mathrsfs}%
\usepackage[title]{appendix}%
\usepackage{xcolor}%
\usepackage{textcomp}%
\usepackage{manyfoot}%
\usepackage{booktabs}%
\usepackage{algorithm}%
\usepackage{algorithmicx}%
\usepackage{algpseudocode}%
\usepackage{listings}%

\usepackage{amssymb}
\theoremstyle{thmstyleone}%
\theoremstyle{thmstyletwo}%

\theoremstyle{thmstylethree}%

\begin{document}

\title[Article Title]{DF-DM: A foundational process model for multimodal data fusion in the artificial intelligence era}


\author*[1,2]{\fnm{David} \sur{Restrepo}}\email{davidres@mit.edu}\equalcont{These authors contributed equally to this work.}

\author[3]{\fnm{Chenwei} \sur{Wu}}\email{chenweiw@umich.edu}
\equalcont{These authors contributed equally to this work.}

\author[4]{\fnm{Constanza} \sur{Vásquez-Venegas}}\email{covasquezv@inf.udec.cl}

\author[1,5]{\fnm{Luis Filipe} \sur{Nakayama}}\email{luisnaka@mit.edu}

\author[1, 6, 7]{\fnm{Leo Anthony} \sur{Celi}}\email{lceli@mit.edu}

\author[2]{\fnm{Diego M} \sur{López}}\email{dmlopez@unicauca.edu.co}

\affil[1]{\orgdiv{Laboratory for Computational Physiology}, \orgname{Massachusetts Institute of Technology}, \orgaddress{\city{Cambridge}, \state{Massachusetts}, \country{United States of America}}}

\affil[2]{\orgdiv{Departamento de Telemática}, \orgname{Universidad del Cauca}, \orgaddress{\city{Popayán}, \state{Cauca}, \country{Colombia}}}

\affil[3]{\orgdiv{Department of Electrical Engineering and Computer Science}, \orgname{University of Michigan}, \orgaddress{ \city{Ann Arbor}, \state{Michigan}, \country{United States of America}}}

\affil[4]{\orgdiv{Scientific Image Analysis Lab}, \orgname{Universidad de Chile}, \orgaddress{ \city{Santiago}, \state{Santiago}, \country{Chile}}}

\affil[5]{\orgdiv{Department of Ophthalmology}, \orgname{São Paulo Federal University}, \orgaddress{ \city{São Paulo}, \state{São Paulo}, \country{Brazil}}}

\affil[6]{\orgdiv{Department of Biostatistics}, \orgname{Harvard TH Chan School of Public Health}, \orgaddress{ \city{Boston}, \state{Massachusetts}, \country{United States of America}}}

\affil[7]{\orgdiv{Department of Medicine}, \orgname{Beth Israel Deaconess Medical Center}, \orgaddress{ \city{Boston}, \state{Massachusetts}, \country{United States of America}}}


\abstract{In the big data era, integrating diverse data modalities poses significant challenges, particularly in complex fields like healthcare. This paper introduces a new process model for multimodal Data Fusion for Data Mining, integrating embeddings and the Cross-Industry Standard Process for Data Mining with the existing Data Fusion Information Group model. Our model aims to decrease computational costs, complexity, and bias while improving efficiency and reliability. We also propose "disentangled dense fusion," a novel embedding fusion method designed to optimize mutual information and facilitate dense inter-modality feature interaction, thereby minimizing redundant information.

We demonstrate the model's efficacy through three use cases: predicting diabetic retinopathy using retinal images and patient metadata, domestic violence prediction employing satellite imagery, internet, and census data, and identifying clinical and demographic features from radiography images and clinical notes. The model achieved a Macro F1 score of 0.92 in diabetic retinopathy prediction, an R-squared of 0.854 and sMAPE of 24.868 in domestic violence prediction, and a macro AUC of 0.92 and 0.99 for disease prediction and sex classification, respectively, in radiological analysis.

These results underscore the Data Fusion for Data Mining model's potential to significantly impact multimodal data processing, promoting its adoption in diverse, resource-constrained settings.}




\maketitle

\section{Introduction}\label{sec1}

In this era, vast amounts of data and information are generated in different settings, fields, formats, and modalities. According to the International Data Corporation (IDC), 7.5ZB are expected to be generated every year \cite{1}, and the ability to process this information and extract knowledge has never been more important \cite{2, 3}. This data can come from diverse sources such as wearable devices, laboratory tests, Electronic Health Records (EHR), among many others, and modalities, such as medical images, clinical notes, and vital signs. The integration of data from different modalities is on the rise in Artificial Intelligence (AI) fields, such as Machine Learning (ML) and Deep Learning (DL), known as multimodal data fusion. This research area, with applications integrating areas from natural language processing to computer vision and beyond, has driven applications in healthcare \cite{4, 5}, integrating EHRs with medical images \cite{6, c.1, c.4, 7, 8, c.6, c.10, 9} or signals from wearable devices \cite{10, 11}. Multimodal fusion has also led to applications such as autonomous driving \cite{12, 13}, environmental sciences applications combining different sensors and satellite data\cite{14, 15}, as well as many other Internet of Things (IoT) applications and system improvements\cite{16, c.8, 17, c.5, 18}.

Recent advances in the DL field, mainly driven by the proposal of the transformer architecture \cite{19}, have seen breakthroughs in the analysis of different modalities, such as text with models like BERT \cite{20} or GPT \cite{21} or images with models like Vision Transformers \cite{22}. These models have been a revolution also in specific fields such as medicine with impressive performances in medical applications such as medical tests or diagnosis \cite{23, 24, 25}. However, some of the limitations of working with these data modalities and these specialized DL models are, the high dimensionality of the data, the heterogeneity of the diverse data modalities and formats, and the amount of data and computational resources required to train robust models for these specific modalities \cite{26, 27, c.7}. Embeddings are low-dimensional numerical vectors learned during DL model training that preserve the original data's most relevant features. Embeddings provide a common format that can solve data extraction and model development in environments where computational resources are scarce. Furthermore, their simplicity gives embeddings a possible way to unify the different data modalities and an efficient and generalizable solution for building multimodal models. 

Since embedding is proposed as a potential solution to data heterogeneity, and computational resource constraints, the DL models used to extract these embeddings must be extremely robust models trained on large amounts of data. In this context, we discuss open-source foundation models, such as Dino v2 \cite{28} for general images or Llama 2 \cite{29} for general text. Foundation models may also be available for specific tasks such as clinical notes with examples like Med-Bert \cite{30}, Bio-Bert \cite{31}, or medical images like Retfound for retinal images \cite{32}.

However, having data and making a prediction is not enough and can cause harm if not done correctly. Adopting robust and reproducible practice is the most important characteristic of any data-related process \cite{33}. Using AI techniques irresponsibly will lead to poor model performance in bias \cite{34, 35, 36, c.2}, robustness, and fairness \cite{37, 38, opendata2}. Acknowledging and addressing potential biases inherent in the data sources and models employed is pivotal. 

There are already previous frameworks for data fusion, starting with the best-known Joint Directors of Laboratories (JDL) \cite{39} and its multiple updates that led to the Data Fusion Information Group (DFIG) \cite{40} model with its most recent version that integrates AI techniques \cite{41}. However, these models do not leverage foundation models and DL nor incorporate best practices for reducing bias and improving the model's robustness, and have a focus only on applications such as industry and the military, leaving out use cases in environments such as healthcare.

In this paper, we delve into a foundational approach for multimodal data fusion, centering the framework on the use of foundation models, vector embeddings, and good practices for data mining. Our work draws inspiration from established frameworks such as the JDL \cite{39} and DFIG \cite{40}. We emphasize the role of embeddings in providing flexibility and improving efficiency, robustness, and fairness, which is even more crucial in low-resource settings where errors and bias are relatively more impactful \cite{42}. We also showed the model's performance in three healthcare use cases: diabetic retinopathy diagnosis using fundus retinal images and EHR data, domestic violence prediction using open data from satellite images and internet data such as online news, and medical diagnosis and gender extraction using radiological images and clinical notes.

To succinctly encapsulate the contributions of this paper, we propose the following bullet points that highlight the primary advancements and demonstrations provided by our research:

\begin{itemize}

\item \textbf{Introduction of a Novel Data Fusion Model:} We have developed an innovative process model for multimodal data fusion, rooted in the principles of foundation models and embeddings. This model is designed to efficiently address the challenges of big data's high dimensionality and heterogeneity, making it particularly suitable for complex environments like healthcare.

\item \textbf{Disentangled Dense Fusion Method:} Our work introduces a deep fusion alignment method that leverages mutual multimodal embedding information. This technique decouples entangled multimodal pairs into compact distinct components: modality-common features and modality-specific knowledge features, reducing the inter-modal redundancy while keeping expressiveness of modality-specific information. We also combine our mutual information decomposition with dense fusion to capture richer modality interactions.

\item \textbf{Three Demonstrative Use Cases:} The efficacy and versatility of our process model and embedding alignment method are showcased through three distinct use cases. These include the prediction of diabetic retinopathy using retinal fundus images and patient metadata, domestic violence prediction through the fusion of satellite images, internet data, and census data, and the detection of clinical and demographic characteristics via radiography images and clinical notes. 
\end{itemize}

\section{Background}\label{sec2}

Data fusion models and frameworks have historically been a great field of research due to the need to take advantage of all the available information. The first model to be proposed and the most widely used is the data fusion model proposed by the JDL, which was introduced in 1987 \cite{43}. The JDL model, mainly created for multi-sensor applications, defined data fusion as combining data to refine state estimates and predictions from a low level (sensor data) to a high level (processes). Subsequent works, such as the revision of the JDL model in 1999 \cite{44}, formalized and refined the JDL model by defining a 0-4 levels model, being the labels: Level 0 - Data Assessment, Level 1 - Object Assessment, level 2 - Situation Assessment, Level 3 - Impact Assessment, and Level 4 - Process Refinement.

Some improvements were proposed during the following years, such as introducing adaptive generalized closed-loop data fusion processes and the perceptual reasoning system \cite{45}. Eventually, a second revision of the JDL model (JDL II) to align with Department of Defense (DoD) priorities, improve quality control, reliability, and consistency, and empathize with the need for co-processing of inferencing processes was proposed in 2004 \cite{46}. Also in that year, due to the shortcomings of the model, mainly related to the lack of clarity of human participation and feedback, the DFIG model was created, adding two levels (Level 5 - User Refinement, Level 6 - Mission Management) to the JDL model [40]. 
However, other adaptations and updates were proposed to the JDL model during the following years, mainly focusing on the optimization and reliability of the model by proposing communication across the JDL levels \cite{47} or implementing different techniques such as game theory \cite{48, 49}, fuzzy logic \cite{50}, ontologies \cite{51} and other mathematical modeling approaches \cite{52, 53, c.9}. Other revisions focused on the fusion, impact, and communication between hard data (machine or environmental data) and soft data (human data). Changes in specific levels such as Levels 1 and 2 incorporating the human in the fusion and preprocessing levels \cite{54}, subsequently in Levels 4 and 5 through hybrid-sensing and hybrid-cognition \cite{55}, or even the incorporating a management network in parallel to the JDL model \cite{56, 57} were proposed.

Finally, in response to the popularity and flexibility of new techniques impulsed by AI and ML, two papers proposed updates to the JDL and DFGI models. An adaptation of the JDL proposing using conversational agents to improve the human-machine interface at Level 5 (User Refinement) was proposed \cite{58}. Subsequently, an adaptation of the DFIG model to incorporate AI, ML, and DL techniques was also presented \cite{41}. The study suggests using ML and DL techniques to model data from different Levels 2 and 3 modalities. The model also proposes using Active Learning (AL) to get human feedback in Levels 5 and 6 and Reinforcement Learning (RL) to improve the system in Level 4. This is the most up-to-date model and will be used as a reference for our proposed model.

The proposed Data Fusion for Data Mining (DF-DM) model is a foundational process model for multimodal data fusion based on DFGI and CRISP-DM. The following sections will introduce the most up-to-date data fusion model (DFGI) with the adaptation for ML and AI \cite{41}, integrate the Cross-Industry Standard Process for Data Mining (CRISP-DM) process model \cite{59}, and present some modifications to implement foundation models and embeddings in the DFGI model. Finally we’ll provide three use cases to show the flexibility and adaptability of our model, as well as usability.

\subsection{Current Data Fusion Model: DFGI Model}\label{subsec21}

\begin{figure}[h]
\centering
\includegraphics[width=1\textwidth]{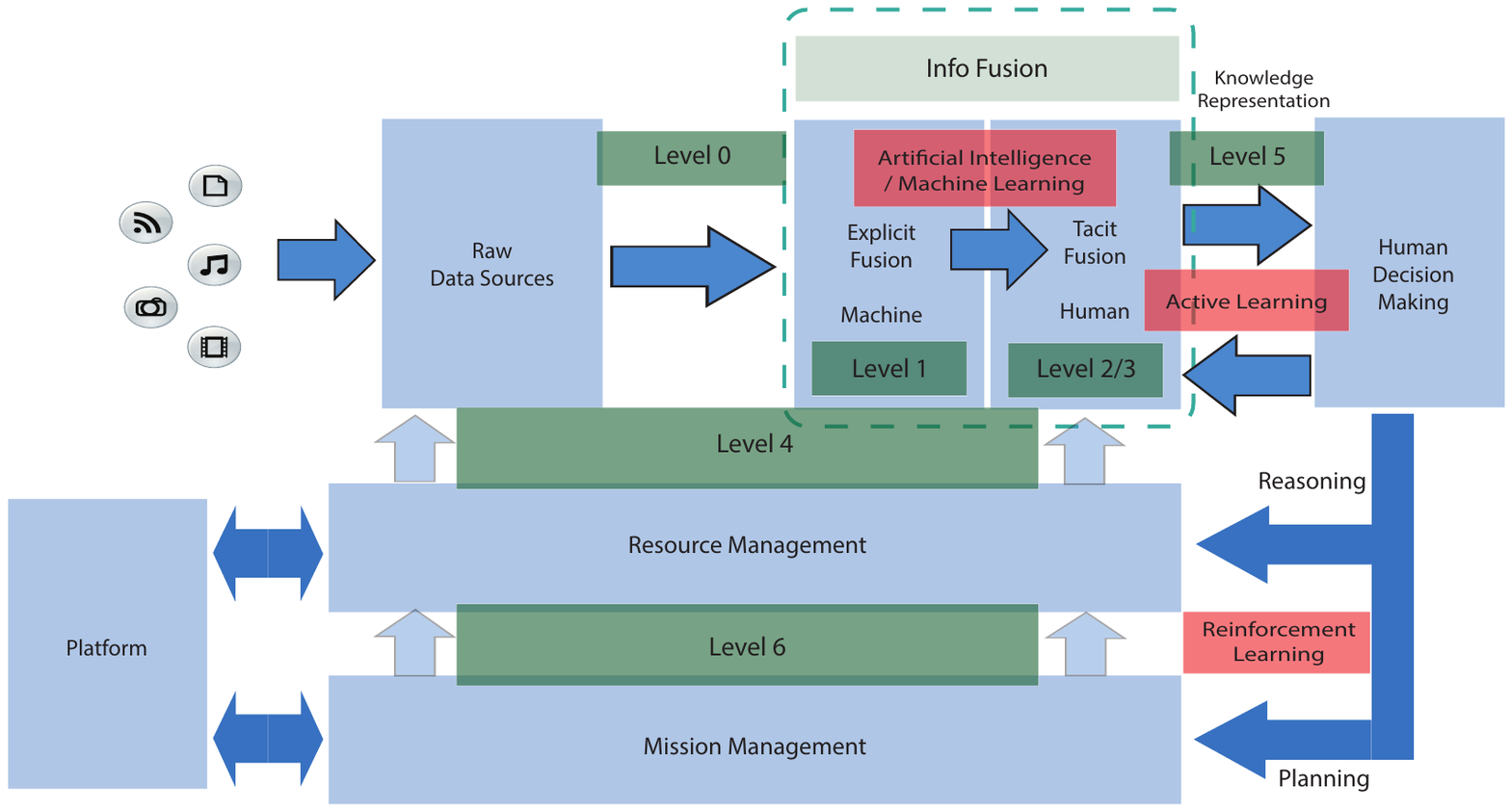}
\caption{DFGI Data Fusion model proposed, including AI and ML from \cite{41}. The Levels where AI and ML techniques are proposed can be seen in red. The original DFGI model can be seen in blue.}\label{fig1}
\end{figure}

The most up-to-date data fusion model is the one proposed by the DFIG \cite{40} and its AI and ML update \cite{41}. This model comprises seven levels, from 0 to 6, that offer a comprehensive framework for data fusion while incorporating AI and ML techniques at different levels, as seen in Figure \ref{fig1}. To understand this model, let’s analyze each one of the levels with an example of data fusion in the healthcare domain:

\begin{itemize}
\item \textbf{Level 0 - Data Assessment:} This level is primarily where data acquisition and preprocessing are. For example, this level can involve collecting various patient data, including electronic health records (EHRs), medical images, wearable device data, and lab reports and their preparation.
\item \textbf{Level 1 - Object Assessment:} It centers on estimating the states of individual parameters and data types. AI and ML techniques are proposed at this level due to their role in predictive tasks such as classification and regression. At this level, the model estimates and predicts the states of individual health parameters, such as vital signs, medical conditions, and medication history. At this point, there is still no data fusion, only individual analysis of each modality/data source.
\item \textbf{Level 2 - Situation Assessment:} At this stage, the model predicts relationships between entities. At this point, data from different modalities and/or sources is finally fused. An example is the measurement of the impact of medications on vital signs or the correlation between lab results and disease progression. 
\item \textbf{Level 3 - Impact Assessment:} Level 3 estimates the effects of planned or estimated actions on situations. We can, for example, measure the impact of different medical interventions on the patient's health. Performance evaluation techniques are used here.
\item \textbf{Level 4 - Process Refinement:} This level focuses on enhancing data collection and processing through adaptive data acquisition and processing. At this level, Reinforcement Learning is proposed to improve the model automatically.  For example, sensor management can collect data from wearable devices to support real-time patient monitoring and intervention.
\item \textbf{Level 5 - User Refinement:} Level 5 involves adaptively determining who can access information, aiding in cognitive decision-making and actions, especially in human-computer interfaces. AL is proposed to introduce humans into the loop at this level. For example, it can support a clinician in making informed decisions by providing relevant patient diagnoses and predictions.
\item \textbf{Level 6 - Mission Management:} Level 6 focuses on the adaptive control of resources to support decision-making. The example involves the control of healthcare resources, such as hospital beds, operating rooms, and healthcare personnel.

\end{itemize}

\subsection{The Integration of the Cross-Industry Standard Process for Data Mining (CRISP-DM)}\label{subsec22}

CRISP-DM was selected to support the DF-DM model because it is considered an industry standard process to guide machine learning and data mining projects \cite{33, 60}. The CRISP-DM model \cite{59} consists of 6 phases:
\begin{enumerate}
    \item \textbf{Business Understanding:} Define the problem, objectives, and constraints from a business perspective.
    \item \textbf{Data Understanding:} Collect and explore the data, gaining insights into its quality and characteristics.
    \item \textbf{Data Preparation:} Clean, transform, and preprocess the data to make it suitable for analysis.
    \item \textbf{Modeling:} Select and apply appropriate predictive models.
    \item \textbf{Evaluation:} Assess the model's performance and its alignment with business goals.
    \item \textbf{Deployment:} Integrate the model into the business process and create a plan for monitoring and maintenance.
\end{enumerate}

One of the greatest advantages of the CRISP-DM model is its flexibility, given that it is not a cascade model. The model allows previous phases to be revisited, thus avoiding the loss of time and resources by having to complete an entire cycle to return to a previous phase. The efficacy and flexibility of CRISP-DM has been shown even in medical settings \cite{61, 62}.
Although the DFGI and CRISP-DM models were developed for different purposes, the DFGI model has remained stuck in an architecture that is still too old for cutting-edge technologies such as ML and DL techniques.
Below, we describe some of the shortcomings of the DFGI model and how they can be addressed by taking components from the CRISP-DM model.

\begin{itemize}
\item \textbf{A closed loop model:} One of the areas for improvement in the DFGI model is the need for more opportunities to revisit the immediately preceding level (this relationship only occurs between Levels 3 and 5). This makes it impossible to act quickly on problems presented in the initial levels, such as 0 or 1. This implies a very large loss of resources and time.
\item \textbf{Do we understand the data?:} One of the most important phases of the CRISP-DM model is understanding the data because it is the primary source of information. If the data is not good enough, no matter how powerful our model is, more is needed. Understanding the nature and quality of multimodal data is important.
\item \textbf{Do we understand the problem?:} A missing aspect in the DFGI model is business understanding. The business understanding phase should be an important feature for all the levels of the DFIG model.
\item \textbf{Correlation or Causality:} At Level 3, we propose to use ML techniques to measure impact. Although traditional ML and/or DL techniques can be effective in many situations, to measure the impact, the integration of a causal analysis is proposed to avoid bias.
\item \textbf{Many modalities, many costs:} The use of multimodal data, although one of the most efficient ways to integrate and take advantage of data, also becomes very expensive when discussing high-dimensional data such as text or images. Given this, we propose using newer techniques and models for dimensionality reduction, such as extracting embeddings from foundation models or using foundation models directly for zero-shot learning tasks.
\item \textbf{Reinforcement Learning:} Although Reinforcement Learning is an effective proposal with many applications, it is only effective in military and industrial environments where rapid simulations or experiments can be performed. However, the use of Reinforcement Learning in environments such as health sciences is a method that takes more time and resources, making it impractical.
\end{itemize}

\section{Integrating Data Mining and Data Fusion in the AI era by Introducing the Fusion Model for Data Mining DF-DM}

\begin{figure}[h]
\centering
\includegraphics[width=1\textwidth]{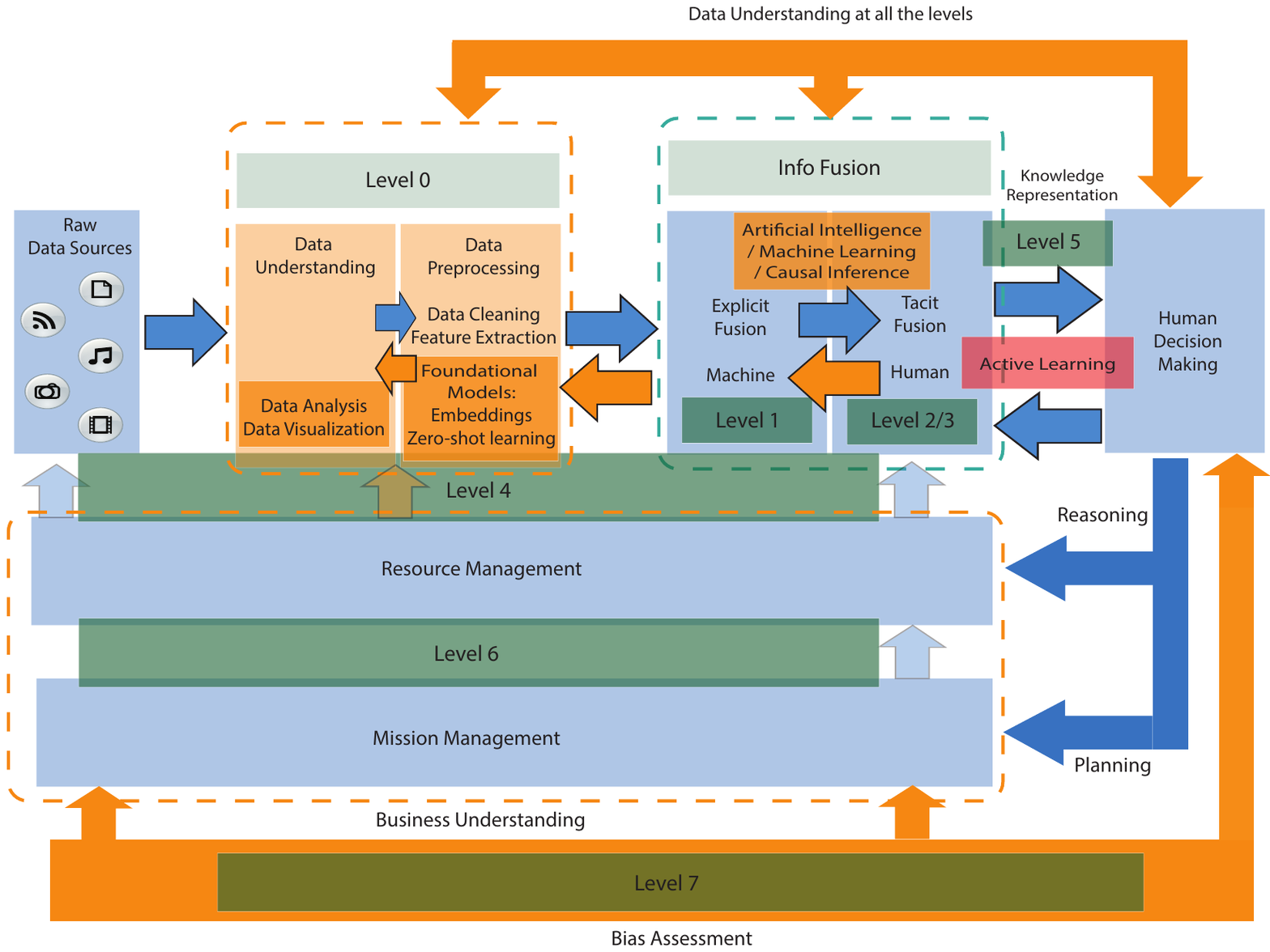}
\caption{The proposed Data Fusion for Data Mining Model (DF-DM). The model is based on the DFGI model, integrating AI and ML but adding other functionalities vital for data mining tasks in orange. }\label{fig2}
\end{figure}

In the era of AI-driven analytics, the integration of data mining and data fusion stands as a paramount challenge, primarily due to the heterogeneity of data and the constraints of computational resources. To address these challenges, we introduce an innovative methodology within our Data Fusion for Data Mining (DF-DM) model (Figure \ref{fig2}), which extends the existing DFGI model by integrating the model with the best practices of the CRISP-DM process model. Additionally our model leverages the use of foundation models and vector embeddings simplifies the processing of diverse data types but also significantly reduces the computational overhead, making advanced data analytics more accessible and efficient.

The rationale behind the DF-DM model's design is twofold. Firstly, by incorporating the CRISP-DM process, we aim to adhere to a proven, structured approach for data mining that emphasizes business understanding, data understanding, bias, and a cyclical process for model refinement. This ensures that our model is not only technically sound but also closely aligned with practical, real-world applications. Secondly, the decision to utilize embeddings and foundation models stems from their ability to reduce high-dimensional data into more manageable, lower-dimensional vectors. This significantly alleviates computational demands and facilitates more effective integration of diverse data types, crucial for addressing the complex challenges of multimodal data fusion in healthcare and beyond.
Before delving into the methodology, it is pertinent to clarify two pivotal concepts:

\begin{itemize}
\item \textbf{foundation models:} foundation models are at the heart of our methodology. These are extensive, pre-trained models that have been developed to capture a wide array of features across specific data modalities, such as text or images. By understanding the general characteristics inherent to each modality, these models can generate useful representations of new, unseen data. This ability is crucial for extracting relevant features without necessitating additional, computationally expensive training phases for each new dataset.

\item \textbf{Embeddings:} Embeddings are also a pivotal role in our methodology, transforming high-dimensional data into a more manageable, lower-dimensional space. Formally, given an input data point $x \in \mathbb{R}^n$, where $n$ is the dimensionality of the raw data in a high-dimensional space. We seek to find its embedding $E \in \mathbb{R}^d$ in a d-dimensional vector space where $d$ is the dimension of the embedding vector and $d << n$. The embedding is generated given the transformation $E=f(x)$, where $f$ denotes the transformation function derived from the foundation model. This process significantly reduces dimensionality, with $d << dim(X)$, facilitating more efficient data analysis. 
\end{itemize}

The proposed model introduces changes to improve the efficiency of the current DFGI model while making the model more reliable and presenting the most recent AI techniques. The Levels offered in the DF-DM model are:

\subsection{Level 0 - Data Assessment:}
This is the lowest level of the model DF-DM. This level consists of 2 steps:

\begin{itemize}
\item \textbf{Data Understanding:} In this step, visualization and assessment of the data quality are essential to creating clear and informative data representations. This section involves selecting appropriate visualization techniques that best convey the patterns and relationships in the data. Pay attention to labeling, scaling, and color choices to ensure the visualizations are easily interpreted. Missing data measures and descriptive statistics are also important.
\item \textbf{Data preprocessing:} Data preprocessing involves cleaning, transformation, normalization, encoding, reduction, and formatting, making it a critical step in the overall DF-DM model. In this step, as shown in Figure \ref{fig3}, using embeddings and foundation models is crucial for generating latent representations and extracting meaningful information from high-dimensional data such as images or text. Techniques such as zero-shot learning or embedding extraction are the key elements in this approach to extract features in low-resource settings. This approach significantly reduces computational costs and storage requirements and creates a unified format that facilitates multimodal data fusion.
\end{itemize}

To facilitate direct communication with Level 5 - User Refinement, we implement a bi-directional interface allowing for real-time feedback and adjustments. This interface could take the form of an interactive dashboard where users can flag issues or suggest refinements, thus enabling a dynamic and responsive system.

This section also includes direct communication with the Level 5 - User Refinement. It allows human users to communicate through a user interface from this initial level, enabling fast error detection in the earliest stages and providing a comprehensive pipeline. For example, if we receive patient data, such as electronic health records (EHRs), medical images, and lab reports. At this stage, we can present information such as keywords or alerts of abnormal values to the clinician. We can also generate text embeddings from the clinical notes in the EHRs, for example, using Llama 2 \cite{29} and image embeddings of the medical image using DINO V2 \cite{28}. The embeddings are vector representations that store all the information and the tabular data in a CSV file.

\subsection{Level 1 - Object Assessment}
\begin{figure}[!h]
\centering
\includegraphics[width=0.9\textwidth]{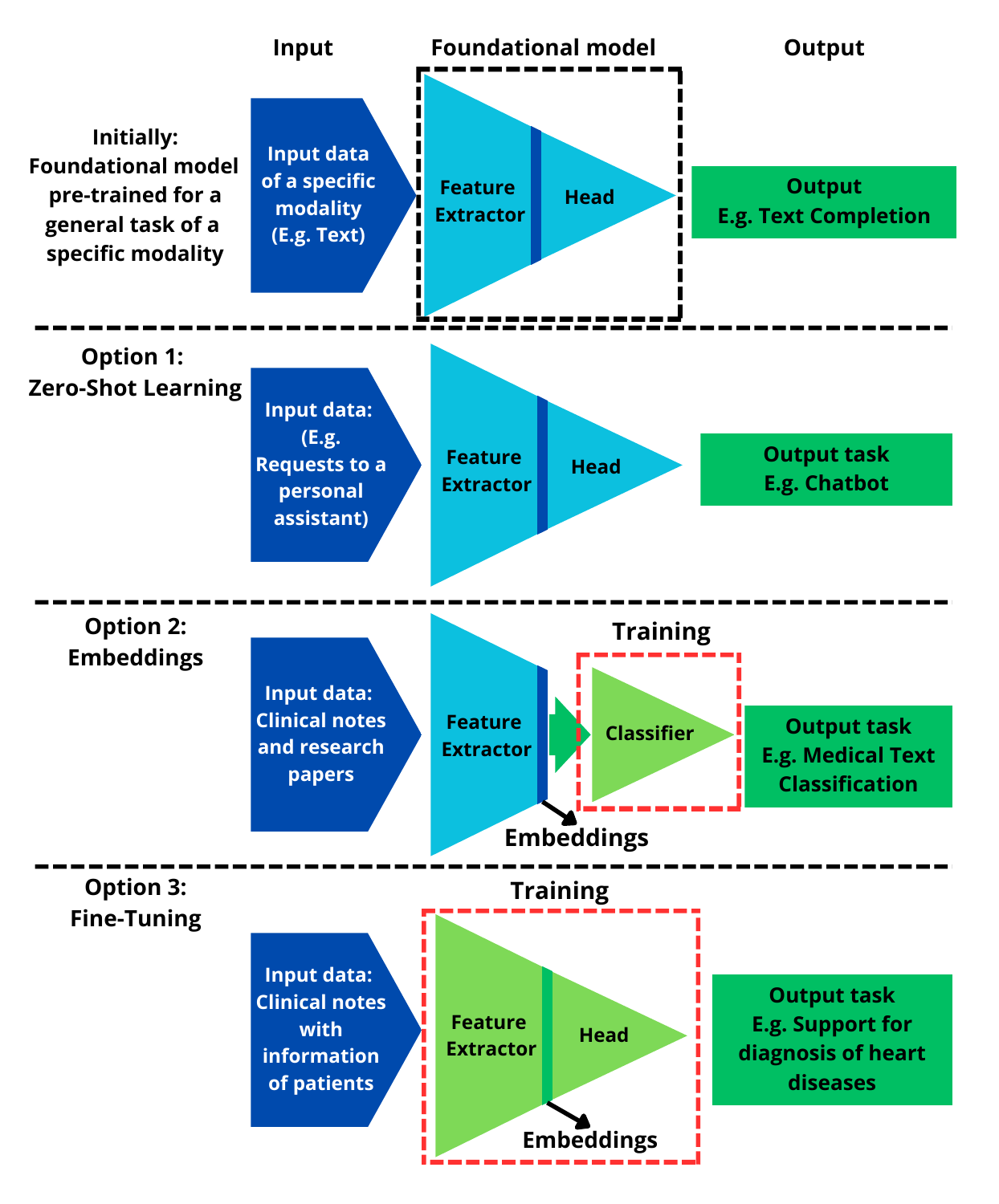}
\caption{Illustrative Framework for Utilizing foundation models in Various Tasks. The figure assumes an initial foundation model for a general task and 3 different options. Option 1 is Zero-shot learning using the foundation model directly for a downstream task. Option 2 suggests the use of embedding, where embeddings of the original data are extracted and used for downstream task training. Option 3 means fine-tuning the full model for a specific task. The resulting model can also be used for embedding extraction.}\label{fig3}
\end{figure}

In Level 1, since the data received from Level 0 are embeddings or features extracted from the foundation models, the data is already in a lower dimensionality and a common format, facilitating different ML and AI techniques. As seen in Figure \ref{fig3}, Option 2, using embeddings, reduces the complexity and dimensionality of the data, making it easy and inexpensive to do the ML modeling because we can use simple models such as linear models. The data resulting from embedding extraction is methodically divided based on data modality (e.g., text, images). This separation enables targeted model training for each data type, enhancing specificity and accuracy. Integration of these segmented models is achieved through a comprehensive analysis framework, allowing for nuanced insights across modalities. For example, using the data from Level 0 in a CSV format due to the embedding extraction, the CSV file can be split for each modality to train a predictive model per modality. ML models can be trained to separately measure the impact of clinical notes, medical images, and tabular data in a given condition.

\subsection{Level 2 - Situation Assessment}
In this level, the modalities can be used to model complex relations across multiple modalities. The outputs of the modalities created in Level 1 can be aggregated in a late fusion approach, or all the features in the CSV file from Level 0 can be concatenated to train a single ML model using an early fusion approach \cite{63}. The choice between late and early fusion approaches is guided by the specific objectives and the nature of the data. Early fusion is preferred when a unified model of all modalities can provide deeper insights, while late fusion is selected for preserving modality-specific characteristics until the final aggregation stage.  For example, using the CSV generated in Level 0 with all the modalities as input of a single ML model will measure the influence of all the modalities together over a specific medical condition.

\subsection{Level 3 - Impact Assessment}
In Level 3, the effect of specific actions or situations in the data is planned and/or estimated. Causal inference can be used to estimate the effect of a particular feature. In this level, for example, causal inference can be used to estimate the effect of a change in a variable, a medication, over a specific outcome.

\subsection{Level 4 - Process Refinement}
This level controls the data retrieval, preprocessing, and analysis strategies based on techniques such as adaptive data acquisition. Adaptive data acquisition is informed by ongoing assessments of data source effectiveness and feedback from Level 5. This includes a systematic evaluation of data quality and relevance, with less valuable sources being phased out or replaced to optimize the data fusion process. At this level, it is important to understand the problem and the business objectives from Level 6 to improve the use of the data, data analysis techniques, and outputs of the data fusion process. At this level, an evaluation of the business understanding and business objectives is performed. For example, Level 5 informs that a data source is generating a lot of noise and wrong measurements, and then Level 4 decides to stop using that variable as part of the data fusion process. This adaptive approach ensures that the data used in the subsequent stages aligns with the desired precision and reliability, contributing to a more effective overall data fusion process.

\subsection{Level 5 - User Refinement}
In this model, the User Refinement is the connection between all the machine levels (0-3) and humans. At this level, specific actions to improve the system can be applied. Here, we communicate with low and high levels to access information and control in each phase. Integration of human insights is facilitated through mechanisms such as feedback loops and supervised learning adjustments, where user inputs directly influence model refinement and data processing strategies. AL is proposed to introduce humans into the machine-learning loop. For example, the data from Level 0 with visualization and statistics can support information to the clinician about which modalities to use or which outcome should be analyzed in Level 2 or 3, making the system more dynamic and robust.

\subsection{Level 6 - Mission Management}
Level 6 focuses on business objectives and business understanding. At this level, we define the main goals of the business and expected outcomes for decision-making. For example, this level involves selecting the specific healthcare workers and sensors that will be used to measure patient data.

\subsection{Level 7 - Bias Assessment}
Bias assessment should influence all the levels of the DF-DM model. Including a dedicated level for Bias Assessment (Level 7) is crucial to address and mitigate potential biases in the multimodal data fusion process. Due to its importance, this level will be explained in a separate section.

\section{Addressing Bias in the DF-DM Model}
While the proposed Data Fusion for Data Mining (DF-DM) model offers significant advantages for multimodal data fusion, it is not immune to the challenges related to bias, which can be present at every level of the data fusion process. As proposed by \cite{64}, bias can manifest in various forms, such as data bias, model bias, or decision-making bias, and it is essential to address these concerns to ensure the reliability and fairness of the model's outcomes.

\subsection{Data Bias}
Data bias can arise from imbalanced datasets or the underrepresentation of certain groups, leading to biased predictions \cite{34, 65, 66, 67}. In healthcare, for example, if the training data predominantly represents a specific demographic group, the model may not perform equally well for other groups. To address data bias in the DF-DM model, we propose the following strategies:
\begin{itemize}
\item \textbf{Data Collection and Sampling:} Ensure that the data collected for the model is diverse and representative of the target population. Stratified sampling techniques can be used to balance data from different groups.
\item \textbf{Bias Detection:} Implement bias detection techniques at different levels of the model to identify and measure bias. Methods like fairness audits and bias indicators should be designed and implemented.
\item \textbf{Data Preprocessing:} Apply preprocessing techniques to mitigate bias, such as re-sampling to ensure fair representation.
\end{itemize} 

\subsection{Model Bias}
Model bias can occur when the model's architecture or training data introduces inherent bias. For example, if the foundation models used for embedding extraction have been trained on biased datasets, this bias can propagate into the DF-DM model. To address model bias, we suggest considering the following:
\begin{itemize}
\item \textbf{Bias Analysis of foundation models:} Before embedding extraction, thoroughly analyze the foundation models to identify and mitigate potential bias. Analyze the current model available and the best option for your use case. Characteristics of the foundation models, such as the source of training data, the original objective of the model, training method, and evaluation metrics, among other characteristics, must be analyzed when choosing the model.
\item \textbf{Causal Inference:} DL approaches are not always enough in all cases. Causal inference, or linear methods as predictive methods, should be used to allow interpretability of the results and relations in the data.
\end{itemize}

\subsection{Decision-Making Bias}
Bias can also emerge in decision-making, especially when human experts are involved. Human decision-makers may unintentionally (or intentionally) introduce bias based on their personal perspectives\cite{68}. To mitigate decision-making bias, we suggest:

\begin{itemize}

\item \textbf{Bias Awareness Training:} Train human decision-makers involved in the process to make them aware of potential biases and encourage fair and objective decision-making.
\item \textbf{Introduce Audits or Peer Review:} Introduce more than one human in the loop. Having more perspectives helps to have a controlled environment and improves transparency and information sharing.
\end{itemize}
We aim to promote fairness, equity, and reliability in multimodal data fusion by addressing these aspects of bias in the DF-DM model. The biases mentioned in this section are not the only types of bias that exist, and each type of bias must be examined independently depending on the case study. We recognize that bias is an ongoing challenge and an active area of research that is nearly impossible to eliminate but should be mitigated.

\section{Use Cases}
To showcase the usefulness of our DF-DM model, we provide three case studies of multimodal data fusion using in three healthcare applications: 

\subsection{Use case 1: Diabetic Retinopathy}
In this use case, we apply the DF-DM model to predict Diabetic Retinopathy (DR) using the open Brazilian ophthalmological dataset (BRSET)\cite{69} leveraging a multimodal approach. BRSET consists of 16,266 fundus images of 8,524 patients, each image, along with demographic and clinical metadata. This section details the exploratory data analysis, preprocessing steps, modality-specific models, the fusion model, and the evaluation of the model's performance and potential bias and limitations. 
To assess diabetic retinopathy classification tasks, a 5-class grouping according to the International Clinic Diabetic Retinopathy (ICDR) classification was conducted. To distinguish between cases requiring closer ophthalmological care, these were grouped into three classes: no diabetic retinopathy, non-proliferative, and proliferative. A division was made into two classes for detecting normal and abnormal classes, indicating the presence or absence of diabetic retinopathy.

\subsubsection{Exploratory Data Analysis and Preprocessing}
The first phase of this use case involved an exploratory data analysis (EDA) of a dataset comprising 16,266 images from Brazilian patients. The EDA focused on assessing the dataset's quality, identifying potential biases, and analyzing the distribution of variables such as gender and age. Key findings include a higher prevalence of female patients (62\%), a significant number of images with missing age data (34\%), and a high imbalance in the county variable, where 100\% of the population in the dataset is from Brazil. The diabetic retinopathy variable presents 5 classes based on the International Clinic Diabetic Retinopathy (ICDR) schema \cite{70}. It is also highly unbalanced, presenting: 15,210 patients with ICDR 0 (94 \%), 162 patients with ICDR 1 (1\%), 310 patients with ICDR 2 (2\%), 190 patients with ICDR 3 (1\%), and 394 patients with ICDR 4 (2\%).

\subsubsection{Preprocessing}
The preprocessing phase was multifaceted:
\begin{itemize}

\item Image Data: Vector embeddings were extracted from images using the Dino V2-Large model \cite{28}, a foundation computer vision model, and stored in a CSV file alongside image IDs.
\item Labels: The dataset's diabetic retinopathy classifications, were restructured into three classification tasks for the analysis given the imbalance: the original 5-class grouping, a 3-class grouping (normal (ICDR score of 0), non-proliferative (ICDR score between 1 and 3), and proliferative ($ICDR score = 1$), and a binary classification (normal vs. diabetic retinopathy ($ICDR score \geq 1$)).
\item Metadata: Variables with over 40\% missing data and variables that can introduce data leakage were dropped. Categorical variables were transformed into a one-hot encoding format. A decision tree classifier was employed to select relevant features, followed by cost complexity pruning. Variables with non-zero feature importance were retained for further analysis (as detailed in Table 1).
\end{itemize}

\begin{table}[h]
\centering
\caption{Metadata Features Selected for Prediction Tasks}
\begin{tabular}{@{}ll@{}}
\toprule
Feature            & Description \\
\midrule
macula             & Macula status (1 for normal and 2 for abnormal) \\
diabetes           & Indicator for self-reported diabetes mellitus (1 for yes and 0 for no) \\
patient\_age       & Age of the patient in years \\
drusens            & Presence of drusens (1 for present and 0 for absent) \\
macular\_edema     & Presence of macular edema (1 for present and 0 for absent) \\
vessels            & Vessel status (1 for normal and 2 for abnormal) \\
camera\_Canon CR   & Indicator for Canon Retinal Camera (1 for yes and 0 for no) \\
myopic\_fundus     & Presence of myopic fundus (1 for present and 0 for absent) \\
camera\_NIKON NF5050 & Indicator for Nikon Retinal Camera (1 for yes and 0 for no) \\
focus              & Focus status (1 for normal and 2 for abnormal) \\
other              & Presence of other abnormalities (1 for present and 0 for absent) \\
amd                & Presence of age-related macular degeneration (1 for present and 0 for absent) \\
patient\_sex       & Gender of the patient (1 for male and 2 for female) \\
scar               & Presence of scars (1 for present and 0 for absent) \\
vascular\_occlusion & Presence of vascular occlusion (1 for present and 0 for absent) \\
\bottomrule
\end{tabular}
\end{table}

\subsubsection{Disentangled Dense Fusion}

\begin{figure}[h]
\centering
\includegraphics[width=1.2\textwidth]{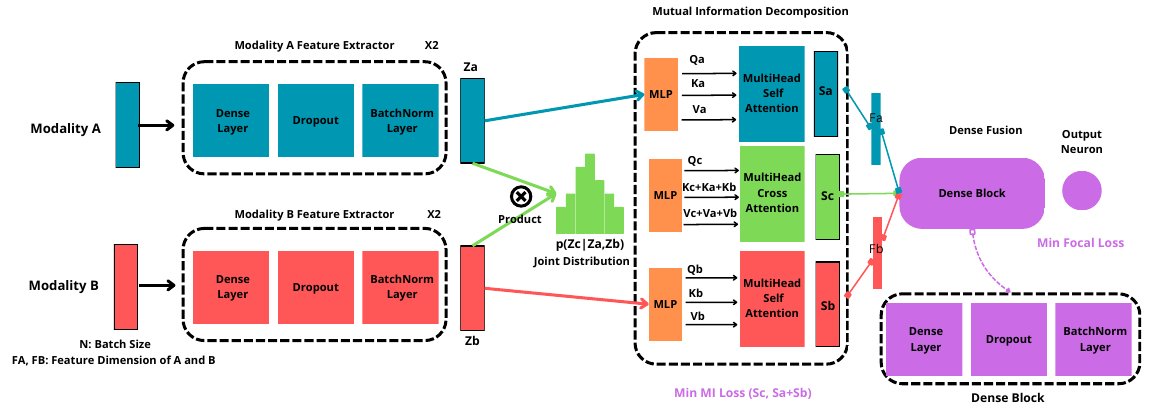}
\caption{Disentangled dense data fusion model for classification tasks.}\label{fig4}
\end{figure}

A significant challenge in fusing information from different types of data, like images and text, is determining how to efficiently combine the overlapping yet critical information that both share, known as the "inter-modal redundancy" issue \cite{71}. This overlap often contains duplicate data that can make it harder to extract useful insights because each type of data (or modality) has its own ratio of useful information to noise \cite{71}. By separating out the unique factors, we can improve the quality of the features we use for analysis while removing unnecessary data. To achieve this, as shown in Figure \ref{fig4}, we utilize a disentangled transformer architecture to decouple the shared and specific representations, eliminating the redundant information and facilitating the fusion model learning. We hope to decompose entangled multimodal data into ideally independent modality-common features $S_c$ and modality-specific features $S_a$, $S_b$. Features from each modality are first multiplied by the Kronecker product to approximate a joint distribution. This layer performs an outer product of the modalities' features,  $C = A \bigotimes B \in \mathbb{R}^{(m\times(a)\times(b))}$, effectively capturing the pairwise interactions between features from different modalities. We apply self-attention to $Z_a$, $Z_b$ to obtain $S_a$, $S_b$, controlling the expressivity of each modality and preventing noisy features. Then we extract the common information of the joint distribution via cross attention of $Q_c$, $K_c+K_a+K_b$, and $V_c+V_a+V_b$ to model modality-common features $S_c$. We minimize the Mutual Information (MI) loss between concatenated $S_a+S_b$ and $S_c$ to preserve modality-specific information. Since the computation of mutual information is intractable, we calculate a variational upper bound called contrastive log-ratio upper bound (vCLUB) [72] as an MI estimator to accomplish MI minimization. Given two variables $a$ and $b$, the $L_vCLUB(a,b)$ is calculated as follows (equation 1)\cite{72}:

\begin{align}
L_v^{CLUB}(a, b) &= \mathbb{E}_p(a, b)\left[\log q_\theta (b|a)\right] - \mathbb{E}_p(a)\mathbb{E}_p(b)\left[\log q_\theta (b|a)\right] \nonumber \\
&= \frac{1}{N^2} \sum_{i=1}^{N} \sum_{j=1}^{N} \left[ \log q_\theta (b_i|a_i) - \log q_\theta (b_j|a_i) \right]
\end{align}

We employ an MLP $q_\theta(b|a)$ to provide a variational approximation of  $q_\theta(b|a)$, so the variational approximation $q_\theta(b|a)$ can be optimized by maximizing the log-likelihood as defined in equation 2:

\begin{equation}
L_{\text{estimator}}(a, b) = \frac{1}{N} \sum_{i=1}^{N} \log q_{\theta}(b_i | a_i)
\end{equation}

Thus the mutual information loss is can be seen in equation 3:

\begin{equation}
    \text{MILOSS} = L_v^{CLUB}(S_a + S_b) + L_{\text{estimator}}(S_a + S_b, S_c)
\end{equation}

Here, after optimizing the mutual information between the modality-specific information and the modality-common information, we utilize dense fusion\cite{73} to allow for denser interaction between modalities. Instead of directly connecting a prediction classifier on top of the fused representation $S_c$, we instead learn deeper representations of the image and non-image features and add skip connections to concatenate with the fused representation to form a final fused embedding, $h_a=f_a (S_a)$ And  $h_b=f_b(S_b)$, where $f_a$ and $f_b$ are fully-connected layers. representation that not only aggregates the modality-specific features, but also incorporates the modality common representation from the previous stage of the network given by the equation 4:

\begin{equation}
    h_{final}=concat(h_a,S_c,h_b)
\end{equation}

Finally, a dense block g is used to generate $y=g(h_{final})$ , and the model is trained by optimizing the prediction loss (focal loss or mean square error loss). This allows for dense interaction of features from each modality, aggregating information across different stages of the network. Finally the final loss defined in equation 5, optimizes a combination of the prediction objective and a mutual information loss controlled by a hyperparameter lambda of value range [0,1].

\begin{equation}
    \text{Loss}_{\text{final}} = L_{\text{objective}}(g(h_{\text{final}})) + \lambda \text{MI}(\text{concat}(S_a, S_b), S_c)
\end{equation}

\subsubsection{Overfitting Analysis}

Separate logistic regression and neural network models were developed for image embeddings and metadata, considering class weights to address overfitting due to data imbalance in logistic regression, and other techniques such as L2 normalization and Early Stopping in the neural network models. The datasets were randomly split into train and test sets, using 70\% for training and the remaining 30\% for testing. The models were evaluated using F1 scores due to the dataset's imbalance (results in Table 2). An early fusion approach was then applied for the fusion model, combining the preprocessed modalities (image embeddings and metadata features) into a single feature set. This feature set was used to train a Logistic regression model using the same train-test split and calculating the same metrics.
Given the big risk of overfitting due to class imbalance, class weights were applied in the loss function in the logistic regression models. Class weights were calculated using the equation described in equation 6, where $W(C)$ indicates the weights for the class $C$, $N$ represents the total data points in the train set, K represents the number of classes, and $N_c$ represents the number of data points of class $C$. 

\begin{equation}
    \alpha(C) = \frac{N}{K \cdot N_c}
\end{equation}

Then the class weights were applied into the class-weighted focal loss, given by equation 7:
\begin{equation}
    FL(p_{t,c}) = - \sum_{c=1}^{C} \alpha_c (1 - p_{t,c})^{y} \log(p_{t,c})
\end{equation}

Where $FL$ is the focal loss for class $c$ given a prediction, $C$ is the total number of classes, $\gamma$ is the focusing parameter, $\alpha_c$ is the class weight.

For the neural network models to avoid overfitting, $L2$ regularization was applied. $L2$ regularization, also known as weight decay, prevents overfitting by penalizing large weights in a model's parameters. This is achieved by adding a regularization term to the model's loss function. The goal of $L2$ regularization is to encourage the model to learn simpler patterns that generalize better to unseen data. The mathematical formulation of $L2$ regularization can be described as in equation 8. Where $L_reg$ indicates regularized loss, $L$ indicates the original loss function, $\theta$ is the weight vector, and $\lambda$ is the regularization strength.

\begin{equation}
L_{\text{reg}}(\theta) = L(\theta) + \frac{\lambda}{2} \|\theta\|_2^2
\end{equation}

\subsection{Results}
We compared the results with the current state of the art methods in BRSET \cite{74, 75}, and with the current state of the art foundation model RetFound \cite{32} to predict diabetic retinopathy. The models demonstrated varying degrees of accuracy and F1 scores across different classification tasks (5-class, 3-class, and binary). The results can be seen in Table 2. Notably, the fusion model outperformed individual modality models, and current state of the art results underscoring the effectiveness of multimodal data fusion in medical diagnoses and the efficacy of our approach.

\begin{table}[!h]
\centering
\caption{Model Results for Different Prediction Tasks for 5 class, 3 class, and 2 class classification}
\begin{tabular}{@{}lllccc@{}}
\toprule
Modality & Model & Accuracy & F1 Macro Avg & F1 Weighted Avg & \\
\midrule
\multicolumn{6}{c}{5 Class - Diabetic Retinopathy} \\
\midrule
Metadata & Logistic Regression & 0.83 & 0.37 & 0.88 \\
 & Neural Network & 0.95 & 0.35 & 0.94 \\
Image Embeddings & Logistic Regression & 0.80 & 0.43 & 0.86 \\
 & Neural Network & 0.96 & 0.52 & 0.95 \\
Raw Images & RetFound \cite{32} & 0.96 & 0.48 & 0.95 \\
Fusion & Logistic Regression & 0.82 & 0.43 & 0.87 \\
 & Dense Fusion & 0.93 & 0.56 & 0.92 \\
 & Disentangled Dense Fusion (Ours) & \textbf{0.97} & \textbf{0.60} & \textbf{0.96} \\
\midrule
\multicolumn{6}{c}{3 Class - Diabetic Retinopathy} \\
\midrule
Metadata & Logistic Regression & 0.89 & 0.56 & 0.91 \\
 & Neural Network & 0.96 & 0.61 & 0.96 \\
Image Embeddings & Logistic Regression & 0.89 & 0.65 & 0.91 \\
 & Neural Network & 0.96 & 0.75 & 0.96 \\
Raw Images & RetFound \cite{32} & 0.96 & 0.78 & 0.96 \\
Fusion & Logistic Regression & 0.90 & 0.67 & 0.92 \\
 & Dense Fusion & 0.93 & 0.73 & 0.93 \\
 & Disentangled Dense Fusion (Ours) & \textbf{0.96} & \textbf{0.79} & \textbf{0.96} \\
\midrule
\multicolumn{6}{c}{2 Class - Diabetic Retinopathy} \\
\midrule
Metadata & Logistic Regression & 0.94 & 0.82 & 0.95 \\
 & Neural Network & \textbf{0.98} & \textbf{0.92} & \textbf{0.98} \\
Image Embeddings & Logistic Regression & 0.93 & 0.78 & 0.94 \\
 & Neural Network & 0.97 & 0.86 & 0.97 \\
Raw Images & ResNet 50 \cite{75} & 0.97 & 0.82 & N/A \\
Fusion & Logistic Regression & 0.97 & 0.87 & 0.97 \\
 & Dense Fusion & 0.98 & 0.91 & 0.98 \\
 & Disentangled Dense Fusion (Ours) & \textbf{0.98} & \textbf{0.92} & \textbf{0.98} \\
\bottomrule
\end{tabular}
\end{table}

\subsubsection{Potential Biases and Limitations}
Despite the promising results, several biases and limitations were noted:
\begin{itemize} 

\item Geographical Bias: The dataset solely comprised Brazilian patients, potentially limiting the model's applicability to other populations.
\item Gender Imbalance: The underrepresentation of male patients (38\%) may result in gender-biased predictions.
\item Age Data: A significant portion of the dataset had missing age data, which might impact the model's effectiveness in age-related analysis.
\end{itemize}

\subsection{Use case 2: Domestic Violence Prediction Using Open Data}

This study leverages the unique capabilities of satellite imagery and internet data to enhance predictive modeling. Satellite images serve as a rich source of information, offering insights into the social determinants of health by revealing spatial and temporal changes in communities and environments \cite{76, 77, 78}. These images can provide valuable indicators related to urban development \cite{79, 80}, population density \cite{81}, and environmental conditions \cite{82, 83}, all of which are crucial in understanding social dynamics that might influence domestic violence. On the other hand, internet data, encompassing Google Trends and online news, acts as a real-time reflection of societal interests and concerns. Internet data has been used to predict the behavior of diseases and social issues \cite{84, 85, 86}, offering a dynamic and current perspective on public discourse and awareness. These alternative data sources are particularly beneficial in low-resource settings where traditional data collection methods might be challenging or insufficient.

\subsubsection{Dataset Description and Data Collection}

This use case employs a novel approach to predict domestic violence in the top 10 Colombian cities (Medellín, Cali, Soacha, Villavicencio, Pasto, Barranquilla, Bucaramanga, Ibagué, Popayán, Cúcuta) with the highest reports of such cases. The dataset spans from January 2016 to January 2023, with the following features:
\begin{itemize}[label={}]
    \item \textbf{Census Data (2018 Colombian Census):} Incorporates social determinants of health and demographic data, providing a comprehensive backdrop of the societal context. The census data included demographic information of the region analyzed. The demographic information included distributions of age, distributions of sex, and ethnicity of the population, while socioeconomic information included information such as access to water, incomes, and level of education.
    \item \textbf{Satellite Images (July 2015 - December 2022):} These images, obtained from Sentinel-2 via Sentinel Hub, cover the targeted cities, offering a spatial perspective.
    \item \textbf{Internet Data:}
    \begin{itemize}
        \item \textbf{Google Trends:} Extracted using the topic "Violence" and keywords related to gender violence in Spanish, offering insights into public interest and awareness.
        \item \textbf{Online Local Newspapers:} Aggregated violence-related news utilizing Media Cloud, capturing media attention and public discourse.
        \item \textbf{GDELT Events:} Focused on relevant events in the selected municipalities, categorized by specific event IDs related to protests and demonstrations for rights and equality.
    \end{itemize}
\end{itemize}

\subsubsection{Exploratory Data Analysis}
An individual data analysis of each modality measuring frequency, format, and missing data, among other variables, was conducted. Exploratory analysis led to the selection of a cohort from Epiweek 51 of 2017 to Epiweek 52 of 2022, primarily due to the lack of comprehensive satellite imagery before 2018.

\subsubsection{Preprocessing}
\begin{itemize}
\item Labels: Represented as the count of domestic violence cases per Epiweek. Epiweeks with no purports were assumed to be 0.
\item Google Trends: Mapped to epiweeks for temporal alignment.
\item Media Cloud \& GDELT: Aggregated by epiweek and city. For GDELT, web scraping supplemented the data with extracted text from URLs. Text data was later discarded due to noise, focusing on quantitative features like publication counts and trend data.
\item Satellite Images: The embeddings of the satellite images were extracted using Variational Autoencoder (VAE) with a ResNet 50 V2  backbone (Fig. 1). This model was trained on a wider set of 81 Colombian cities’ images from 2016 to 2018. The cohort of 81 Colombian cities was chosen to expand the dataset’s diversity of landscapes by incorporating 81 cities spatially distributed throughout the country. This spatial and temporal shift allows us to avoid data leakage, incorporating at the same time a greater diversity of spatial environments that allow the model to extract better quality embeddings. Then, the embeddings were extracted from the latent space for dimensionality reduction. The VAE was used to sample all the images into a lower dimensional space, sampling to a normal distribution. The Variational Autoencoder used for processing satellite images is fundamentally based on two mathematical components: reconstruction loss and Kullback-Leibler (KL) divergence. The reconstruction loss was, in this case, the mean squared error mse, as can be seen in equation 9. The KL Divergence measures how much the learned distribution $q(z | x)$ (the encoder's output) deviates from the prior distribution $p(z)$, which is often assumed to be a standard normal distribution. The KL divergence is given by equation 10. where $\mu$ and $\sigma$ are the mean and standard deviation of the learned distribution, and $J$ is the dimensionality of the latent space. Finally, the total loss is just the sum equation 11. 
In cases like satellite images, embeddings generated by foundation models such as DINO V2 may be less effective. In this case, we generated our embeddings using a VAE and compared them with those generated using DINO V2. The two methods were evaluated in predicting domestic violence. In all cases, the embeddings generated with our approach improved the results of DINO V2, as can be seen in Table 3.
\end{itemize}

\begin{equation}
\text{Reconstruction Loss} = \frac{1}{n} \sum_{i=1}^{n} (x_i - \hat{x}_i)^2
\end{equation}

\begin{equation}
\text{KL Divergence} = -\frac{1}{2} \sum_{j=1}^{J} \left(1 + \log(\sigma_j^2) - \mu_j^2 - \sigma_j^2\right)
\end{equation}

\begin{equation}
\text{Total Loss} = \text{Reconstruction Loss} + \text{KL Divergence}
\end{equation}

\begin{figure}[h]
\centering
\includegraphics[width=0.9\textwidth]{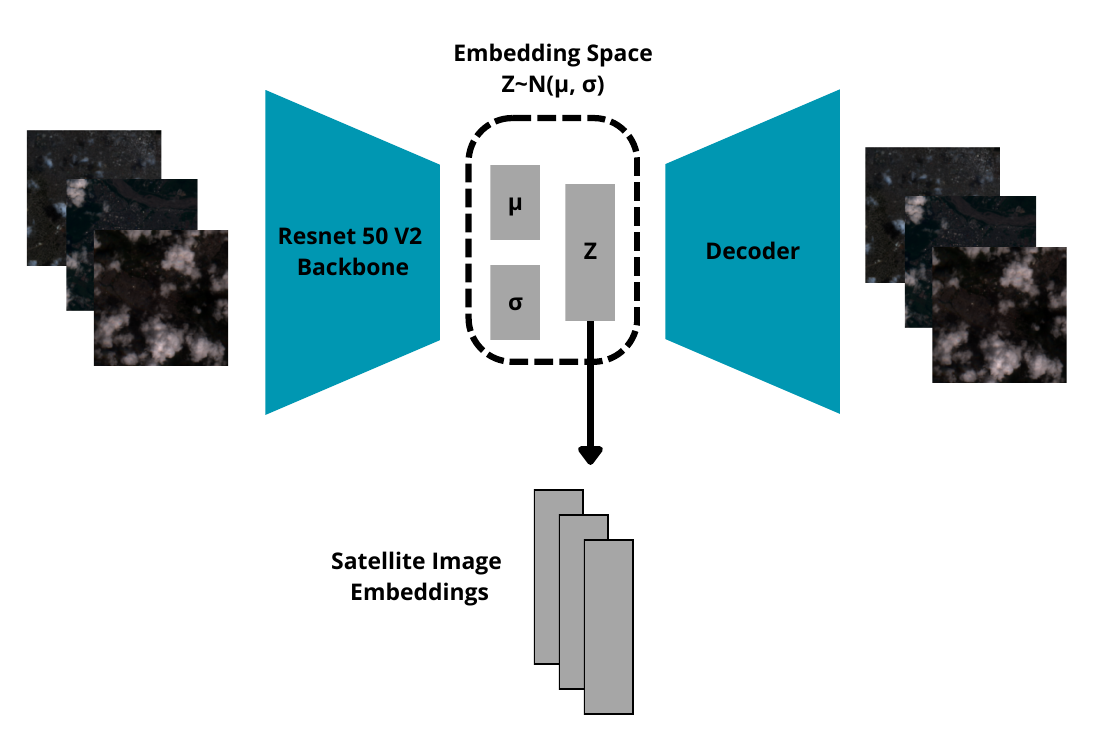}
\caption{Satellite image embedding extraction approach using a variational autoencoder with a Resnet 50 V2 backbone as encoder.}\label{fig5}
\end{figure}

\subsubsection{Disentangled Dense Fusion for Time Series}
\begin{figure}[!h]
\centering
\includegraphics[width=1.1\textwidth]{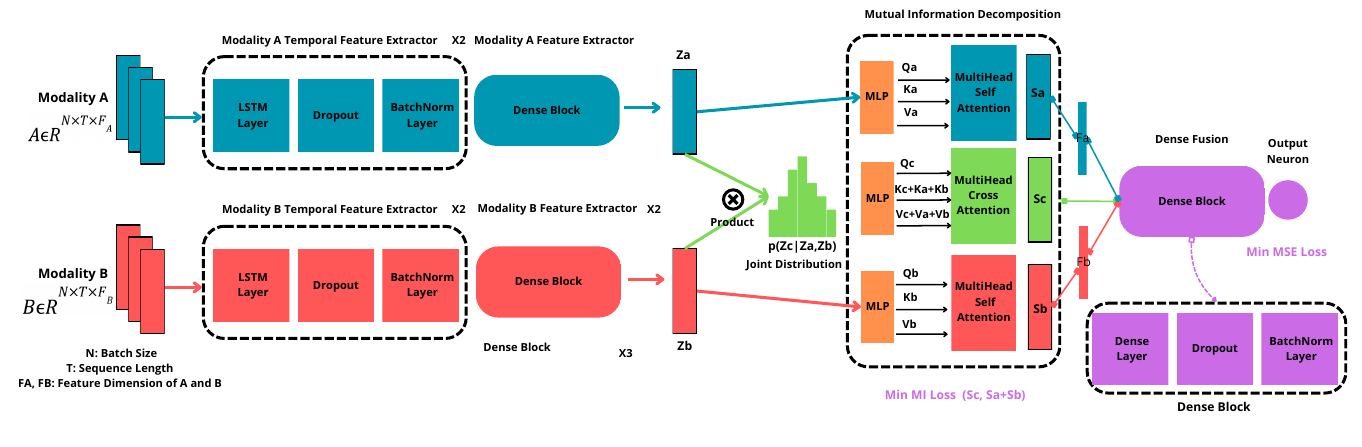}
\caption{Disentangled dense data fusion for temporal prediction tasks.}\label{fig6}
\end{figure}

The predictive modeling was structured as a regression task using a sliding window of 3 epiweeks. The data was chronologically split (80\% train, 20\% test) and normalized. The model architecture can be seen in Figure \ref{fig6}. The model was designed to accept two modalities as input A and B. Each modality is passed through two temporal feature extraction blocks and two modality-specific feature extractor networks. The rest of the architecture is the same as our Disentangled Dense Fusion for classification tasks, except that the objective loss term in the final loss here is mean square error loss instead of the focal loss/cross entropy.

To avoid overfitting during the training of the model, early stopping was applied using an independent evaluation set as we evaluate the output of the model at the end of each training epoch. Since overfitting is caused by a continuous decrease of the training loss and increase in training performance, while a continuous loss in the capacity of the model to generalize; if the performance of the model in the evaluation loss starts decreasing, overfitting is observed. If after a number of epochs, defined by 7 in our case, the performance in validation keeps decreasing, then the model stops the training and we take the weights with the best validation performance for further testing.

\subsubsection{Results}
To evaluate the modalities’ performance in predicting domestic violence, we conducted experiments where we tested different combinations of data sources and  evaluated using metrics such as MAE, MSE, SMAPE, and R2. All the experiments were run 3 times, and the average value and standard deviation were reported, as seen in Table 3. The performance of our embedding generation method was also assessed using the fusion model to combine metadata and image embeddings to predict domestic violence. The same metrics were measured and reported in Table 3. We can see that simply adding more information without regularization will not help the learning, as learning all modalities concatenated together with MLP performs worse than just Internet Data+Census data. Our disentangled dense fusion method indeed eliminates the redundant information across modalities and achieves around 4.2\% $R^2$ improvement over dense fusion.

\begin{table}[!h]
\centering
\caption{Comparison of the 3 modalities (Census data, Internet data and Satellite Images) for domestic violence prediction using our fusion model.}
\label{tab:modality_comparison}
\begin{tabular}{@{}lcccc@{}}
\toprule
Modality & MAE & RMSE & sMAPE & R2 \\
\midrule
Satellite Images (DinoV2) & 27.135±0.535 & 38.287±1.519 & 53.385±5.934 & 0.358±0.050 \\
Satellite Images (Resnet50 v2) & 25.885±4.262 & 33.897±4.219 & 46.034±4.939 & 0.490±0.127 \\
Census Data & 28.039±1.082 & 41.862±5.270 & 46.236±2.870 & 0.221±0.194 \\
Satellite Images + Census Data & 25.885±4.262 & 33.897±4.219 & 46.034±4.939 & 0.490±0.127 \\
Internet Data + Census Data & 22.393±2.345 & 28.766±2.691 & 38.487±2.227 & 0.691±0.057 \\
All modalities (MLP) & 22.487±2.260 & 29.994±2.391 & 37.992±7.380 & 0.6652±0.054 \\
All modalities (Dense Fusion) & 21.548±0.724 & 25.390±0.185 & 25.732±0.580 & 0.812±0.012 \\
All modalities (Ours) & \textbf{14.664±0.652} & \textbf{20.02±0.327} & \textbf{24.868±0.471} & \textbf{0.854±0.015} \\
\bottomrule
\end{tabular}
\end{table}

\subsubsection{Bias Consideration}

Notably, the model exhibits biases:

\begin{itemize} 
\item Geographical Limitation: Data from only 10 cities may not represent broader trends in Colombia.
\item Sampling Bias: The dataset is limited to the top 10 Colombian cities with the most reports of domestic violence. This geographic focus may not capture the nuances and characteristics of domestic violence incidents in smaller cities or rural areas
\item Potential Shortcut Learning: There's a risk that the model learns shortcut features, like seasonal patterns, rather than actual indicators of domestic violence.
\item Internet Data and Public Perception Bias: Using Google Trends and online news articles as part of the features may introduce biases related to public perception and media attention, which do not always accurately reflect the actual incidence of domestic violence.
\end{itemize}

\subsection{Use case 3: Chest X-ray diagnosis and bias identification using MIMIC CXR}

For our third use case, the Medical Information Mart for Intensive Care in its Chest X-ray version (MIMIC-CXR) \cite{87} was used. MIMIC-CXR is a large dataset, publicly available, that contains a collection of chest radiographs paired with clinical notes, offering a perfect multimodal X-ray dataset used in many multimodal tasks. The dataset was collected from the Beth Israel Deaconess Medical Center in Boston, and is composed of 371,920 chest X-ray images from 227,943 imaging studies of 65,079 patients. The dataset includes many tasks including clinical tasks such as disease prediction, and fairness tasks such as race identification, or sex prediction. In this case, we will evaluate our method on the clinical task disease diagnosis, and the fairness task sex prediction.

\subsubsection{Dataset Description and Data Collection}
Given that the focus of this analysis is sex prediction, and disease prediction. In this case, we will center our analysis on those two variables. As a first step we excluded all patients with undetermined race to avoid incomplete data and reports, our cohort contains a subset of 153,128 image and text pairs. The sex prediction contains the categories male and female, and presents a distribution of 82,026 (53.56 \%) male patients, and 71,102 (46.43\%) female patients. The disease prediction was grouped into 4 possible conditions labeled as: others with 90,843 (59.32\%) data points, no finding with 47,184 (30.81 \%) data points, pneumonia with 11,202 (7.31\%) data points, and Lung Lesion with 3,899 (2.55\%) data points.

\subsubsection{Preprocessing}
To prepare the MIMIC-CXR dataset for analysis, we implemented several preprocessing steps:

\begin{itemize}
\item Exclusion of Incomplete Records: Patients with unspecified race were excluded to ensure an analysis of patients with full metadata and reports, reducing the number of data points from 371,920 to 153,128.
\item Image Conversion: All X-ray images were converted to JPG format with a resolution of 224x224 pixels to reduce computational demands without significantly compromising image quality.
\item Train-test split: We splitted the dataset into training, testing, and validation splits using the official split defined in the original dataset.
\item Image Embedding Extraction: We utilized Dino V2 to convert images into vector embeddings, facilitating efficient computational processing and analysis.
\item Text Embedding Extraction: We used the language foundation model Llama 2 with its 7 billion parameter configuration to convert the text in the clinical notes into vector embeddings, enabling a nuanced understanding of clinical notes in a lower dimensional and computationally efficient representation.
\end{itemize}

\subsubsection{Modeling}
We apply the exact same training set up, model architecture and loss optimization as in the diabetic retinopathy classification task, as shown in Figure \ref{fig4}. The single modality models were a logistic regression model, and a simple neural network composed. Early stopping with patience = 7  was applied. 

\subsubsection{Results}
The performance metrics measured for this task were the accuracy and the Area Under the Curve (AUC) with macro average, since these are the most commonly used metrics for this specific dataset. The results of sex prediction were measured using the single modality models, and multimodal models, and compared with related works. 
As can be seen in Table 4, for age prediction, our model archived an AUC of 0.99 for the Fusion model using a neural network archiving same results as the state of the art models in this same task and datasets using larger models like DenseNet-121 \cite{88}, ResNet18, ResNet50, VGG19, or InceptionV3 \cite{89}. 
For the disease classification, results can be seen in Table 5. The best performing model was the logistic regression using only text data reaching a macro AUC of 0.92  and the fusion model using the neural network with an accuracy of 0.80. These results are comparable with related work results for clinical diagnosis using MIMIC CXR, surpassing the macro AUC results reported in literature archiving 0.89 \cite{90, 91}.

\begin{table}[!htbp]
\centering
\caption{MIMIC CXR Results for Sex Classification Task}
\label{tab:sex_classification}
\begin{tabular}{@{}llcc@{}}
\toprule
Modality & Model & Accuracy & AUC Macro Avg \\
\midrule
\multicolumn{4}{l}{Sex Classification} \\
\midrule
Images & Logistic Regression & 0.92 & 0.97 \\
 & Neural Network & 0.92 & 0.98 \\
Text & Logistic Regression & 0.76 & 0.86 \\
 & Neural Network & 0.77 & 0.88 \\
\addlinespace
Fusion & Logistic Regression & 0.92 & 0.97 \\
 & Dense Fusion & \textbf{0.94} & \textbf{0.99} \\
 & Disentangled Dense Fusion (Ours) & \textbf{0.94} & \textbf{0.99} \\
\bottomrule
\end{tabular}
\end{table}

\begin{table}[!htbp]
\centering
\caption{MIMIC CXR Results for Disease Classification Task}
\label{tab:disease_classification}
\begin{tabular}{@{}llcc@{}}
\toprule
Modality & Model & Accuracy & AUC Macro Avg\\
\midrule
\multicolumn{4}{l}{Disease Classification} \\
\midrule
Images & Logistic Regression & 0.44 & 0.71 \\
 & Neural Network & 0.50 & 0.63 \\
Text & Logistic Regression & 0.74 & 0.72 \\
 & Neural Network & 0.78 & 0.76 \\
\addlinespace
Fusion & Logistic Regression & 0.48 & 0.74 \\
 & Dense Fusion & 0.76 & 0.78 \\
 & Disentangled Dense Fusion (Ours) & \textbf{0.80} & \textbf{0.84} \\
\bottomrule
\end{tabular}
\end{table}

\subsubsection{Bias Consideration}

\begin{itemize}
\item Geographical Bias: The MIMIC-CXR dataset is derived from a single hospital in Boston, limiting the model's applicability to populations in different geographic regions with varying disease prevalence and demographic characteristics.
\item Demographic Representation: The inclusion of sex labels allows for the assessment of model performance across different demographic groups. However, it also highlights the risk of perpetuating existing biases in medical diagnosis if not carefully addressed. Other analysis such as fairness analysis on variables like sex and individual group performance should be addressed.
\item Disease Label Bias: The classification of diseases in the dataset may reflect historical diagnostic biases, potentially influencing the AI models' training and predictions.
\end{itemize}
To mitigate these biases, strategies such as diversifying the training data, implementing fairness-aware machine learning techniques, and conducting extensive validation across different populations were recommended.

\section{Discussion}\label{sec12}

The framework proposed in this paper presents a novel approach for multimodal data fusion centered on the use of embeddings, foundation models, and data mining techniques. It addresses an efficient process for extracting knowledge from diverse data modalities while considering resource constraints. In this section, we discuss the advantages and implications of our approach, contrast it with existing models, and discuss potential limitations.

\subsection{Advantages of the Proposed Framework}
\begin{itemize}

\item \textbf{Efficiency and Flexibility:} Our framework introduces a direct and efficient communication channel between the human-in-the-loop and all model levels, enhancing the ability to detect and correct errors at early stages. This feature is vital in resource-constrained settings and real-time decision-making scenarios, such as healthcare or IoT applications.
\item \textbf{Revisitation of Levels:} By allowing revisits to previous levels, the framework accommodates changes in data quality or requirements. This flexibility is a significant departure from the one-way approach of traditional DFGI and JDL models.
\item \textbf{Enhanced Data Understanding:} Incorporating data understanding from the CRISP-DM model into our framework enhances data quality at its earliest stages. Understanding the nature and quality of diverse data modalities becomes a cornerstone for effective data processing, fusion, and modeling reliability.
\item \textbf{Dimensionality Reduction with Embeddings:} Using foundation models and embeddings introduces an efficient solution for high-dimensional modalities like text and images. This method reduces computational costs for model training and data storage, making it a practical choice for resource-constrained environments.
\item \textbf{Causal Inference:} By introducing the possibility of causal inference at Level 3, our framework mitigates the impact of bias, especially when data collection is challenging. This addition provides more accurate and realistic predictions in uncertain environments, such as medical or environmental applications.
\item \textbf{Business Understanding:} Our framework emphasizes the importance of business understanding at the initial stages and throughout the entire process. This holistic approach ensures that the fusion of data aligns with the overarching objectives and constraints of the application.
\item \textbf{Bias mitigation:} Our framework incorporates a dedicated level (Level 7 - Bias Assessment) to systematically address and mitigate bias throughout the entire DF-DM model. By recognizing that bias can manifest at different stages, we provide some strategies for each aspect. This proactive approach involves techniques such as diverse data collection, bias detection, preprocessing, analysis of foundation models, causal inference, bias awareness training, and audits or peer review.
\end{itemize}

\subsection{Comparison with Existing Models}
While traditional data fusion models, such as the DFGI model, offer a structured approach, they often need more flexibility and efficiency to handle diverse data modalities, particularly in resource-limited settings. Our proposed framework bridges these gaps by drawing inspiration from the well-established DFGI model and integrating the flexibility and modern data understanding elements of the CRISP-DM model and the most up-to-date DL techniques. The introduction of foundation models and embeddings further enhances the efficiency and effectiveness of the proposed model.

\subsection{Dense Mutual Information Model}
Our approach systematically maximizes the shared information content between the embeddings of different modalities while allowing for dense modality interactions. This method ensures that the fusion process retains the most relevant and complementary features across data sources, facilitating a more coherent yet diverse representation of the data.
Our approach stands out for its ability to navigate the inherent information optimization challenges of multimodal fusion, such as information redundancy caused by high dimensionality and noisy inputs, and poor joint representation modeling due to heterogeneity of the data sources. Traditional methods often struggle to reconcile differences between modalities and learn useful joint representations, leading to suboptimal fusion outcomes as can be seen in use case 2, which shows that monotonically adding modalities does not guarantee performance boost. In contrast, our model leverages the mutual information metric to identify and disentangle the underlying correlations between modalities, showing improvement across 3 use cases. Furthermore, the model's lightweight and efficient design ensures its applicability in resource-constrained settings, making it a versatile tool for a wide range of applications in healthcare, environmental monitoring, and beyond.

\subsection{Use Cases}

\subsubsection{Use case 1: Diabetic Retinopathy}
In the case of Diabetic Retinopathy, our DF-DM model employed a logistic regression, and a neural network-based approach, demonstrating that a relatively simple machine learning algorithm can yield significant results even outperforming the state of the results in diabetic retinopathy using BRSET. This way, the DF-DM model proves to be particularly beneficial in environments where resources and specialized knowledge are scarce.
In this case, we can also see how the use of a fusion model represented an improvement in the performance of diabetic retinopathy classification. By combining different modalities, the fusion model enhanced the model's performance.

\subsubsection{Use case 2: Domestic Violence Prediction Using Open Data}
In the context of predicting domestic violence, the model showcased its ability to tackle complex social issues using publicly available data. This is particularly relevant in areas where data collection faces economic, geographical, or social barriers. The methodology can be generalized and adapted to various contexts, emphasizing its versatility and applicability.

Notably, the use of self-supervised learning in domain-specific scenarios, such as our Resnet 50 V2 VAE model, outperformed foundation computer vision models like Dino V2. This highlights the importance of domain specific knowledge for data analysis and model training.

The introduction of a temporal data fusion model was fundamental in handling heterogeneous data. However, the model using internet and census data improved the performance of the model employing all modalities. This could be attributed to the potential noise in satellite imagery, which, while informative, has less predictive capacity than internet data for predicting domestic violence. Incorporating attention mechanisms in the fusion process could mitigate this issue.

\subsubsection{Use case 3: MIMIC CXR Clinical notes and x-ray images}

The MIMIC-CXR dataset presented an evaluation of the DF-DM model's efficacy across both clinical and fairness tasks, specifically disease diagnosis and sex prediction. This dual-focused analysis underscores the versatility of our model, not only in tackling complex medical diagnostic challenges but also in addressing critical issues of bias within AI-driven healthcare solutions. The implementation of our model on this extensive dataset, encompassing a broad spectrum of chest radiographs paired with clinical notes, highlights the model's capability to process and derive insights from multimodal data effectively. This ability is paramount in the context of healthcare, where the integration of various data types can significantly enhance diagnostic accuracy and patient outcomes.
By achieving superior, performance metrics against state-of-the-art models in these tasks, our model demonstrates its potential to serve as a valuable tool in clinical settings. The inclusion of sex prediction as a fairness task further enriches our analysis, providing a platform to detect and mitigate potential biases inherent in machine learning models. Such considerations are crucial in ensuring that AI-driven healthcare solutions promote equity and do not perpetuate existing disparities.
The successful application of the DF-DM model in this use case not only validates its conceptual framework but also accentuates its practical utility in real-world scenarios. By navigating the challenges presented by the high-dimensional and heterogeneous nature of the MIMIC-CXR dataset, our model underscores the importance of leveraging embeddings and foundation models to streamline data processing and enhance computational efficiency. 

\subsubsection{General Comments}
Both case studies underline the pivotal role of data analysis in detecting noise, removing outliers, and establishing a robust cohort. This process is essential in all data fusion projects to avoid costly errors later in the project.
The use of embeddings in our model provided a simple and cost-effective solution, facilitating the fusion process. Additionally, acknowledging data bias is crucial in every step of data fusion. Recognizing the limitations of data and modeling approaches is vital to prevent future harm and bias.
Utilizing diverse metrics offered a more comprehensive understanding of the results, reducing the risk of further bias. 
Although the DF-DM process model applied in three use cases showed its effectiveness and potential applicability in other contexts, it is important to note the limitations of our approach. While successful in the specific cases of Diabetic Retinopathy, Domestic Violence Prediction, and Disease and Sex classification from radiological data, applying this model to different use cases might yield other challenges, diverse results, and unique specifications. Each application demands a tailored approach, considering the distinct nature of the data and the specific problem at hand.

\subsection{Limitations and Future Directions}
It is essential to acknowledge that our framework, while addressing many challenges in multimodal data fusion, has limitations. The effectiveness of our approach depends on the availability of foundation models, which may not always be accessible in specific domains. Additionally, the human-in-the-loop approach introduced in Level 5 may require additional resources and time.
As a future direction, research can focus on optimizing the framework for scenarios where foundation models are not readily available, exploring the development of more efficient techniques for dimensionality reduction, and further refining the interaction between human experts and the model to minimize resource requirements.

\section{Conclusion}

In this paper, we introduced the Data Fusion for Data Mining (DF-DM) model, a groundbreaking process model that not only leverages the power of embeddings and foundation models for efficient data fusion but also introduces a novel mutual information based dense fusion model. By optimizing for mutual information, our model ensures that the fused data representation captures the most relevant shared information between modalities while ensuring modality-specific expressiveness, significantly improving the cohesion and efficacy of the fusion process. Our approach enhances efficiency, flexibility, and data understanding while considering resource constraints. By bridging the gaps in existing models, we provide a pathway for improved data fusion in a wide range of applications. We also provided a validation of the model showcasing three healthcare scenarios proving the effectiveness of the approach in healthcare settings.
The proposed DF-DM and mutual embedding information alignment model offer a foundational approach to multimodal data fusion, addressing the challenges of efficiency, flexibility, and data understanding. However, it is essential to acknowledge and actively mitigate biases that may arise in the complex process of integrating diverse data modalities. Including Level 7 - Bias Assessment and the outlined strategies reflect our commitment to promoting fairness, equity, and reliability in the fusion of multimodal data.
As the volume and complexity of multimodal data continue to grow, this framework offers a promising solution that combines the best of traditional data fusion models with the most used techniques in data mining and deep learning with embeddings and foundation models. This research provides a way for more efficient and resource-aware data fusion processes, contributing to advancements in healthcare, environmental sciences, and beyond.

\section*{Acknowledgement}

LAC is funded by the National Institute of Health through R01 EB017205, DS-I Africa U54 TW012043-01 and Bridge2AI OT2OD032701, and the National Science Foundation through ITEST 2148451. DML was funded by a grant from the Colombian Agency of Science, Technology, and Innovation Colciencias under Call 896-2021, project “Knowledge management on the effects of violence on health in the Pacific region of Nariño. A social determinants of health approach”, project ID 110489684405, Contract. 647-2021.



\end{document}